\pdfoutput=1
\documentclass[11pt]{article}

\usepackage[]{acl}

\usepackage{times}
\usepackage{latexsym}
\usepackage{algorithm}
\usepackage{algorithmic}
\usepackage[switch]{lineno} 
\usepackage[utf8]{inputenc}
\usepackage{CJKutf8}
\usepackage{multirow}
\usepackage{booktabs}
\usepackage{graphicx}
\usepackage{amsmath}
\usepackage{xpinyin}
\usepackage{pifont}
\usepackage{subfigure}

\usepackage{xcolor}

\newcommand{\MethodName}{ECOPO}
\DeclareMathOperator*{\argmax}{arg\, max}

\usepackage[T1]{fontenc}
\usepackage[utf8]{inputenc}

\usepackage{microtype}

\title{The Past Mistake is the Future Wisdom: Error-driven Contrastive \\ Probability Optimization for Chinese Spell Checking}

\author{Yinghui Li$^{1}$\thanks{ $^*$ indicates equal contribution. Work is done during Yinghui's internship at Tencent Cloud Xiaowei.},~~Qingyu Zhou$^{2*}$,~Yangning Li$^{1}$,~Zhongli Li$^{2}$,~Ruiyang Liu$^{4}$,\\ ~\textbf{Rongyi Sun}$^{1}$,~\textbf{Zizhen Wang}$^{2}$,~\textbf{Chao Li}$^{2}$,~\textbf{Yunbo Cao}$^{2}$ \and \textbf{Hai-Tao Zheng}$^{1,3}$\thanks{ $^{\dagger}$ Corresponding author: Hai-Tao Zheng. (E-mail: zheng.haitao@sz.tsinghua.edu.cn)} \\
        $^{1}$Tsinghua Shenzhen International Graduate School, Tsinghua University \\ 
        $^{2}$Tencent Cloud Xiaowei, $^{3}$Peng Cheng Laboratory \\
        $^{4}$Department of Computer Science and Technology, Tsinghua University\\
        \texttt{liyinghu20@mails.tsinghua.edu.cn},
        \texttt{qingyuzhou@tencent.com}}

\begin{document}

\maketitle
\begin{abstract}
Chinese Spell Checking (CSC) aims to detect and correct Chinese spelling errors, which are mainly caused by the phonological or visual similarity.
Recently, pre-trained language models (PLMs) promote the progress of CSC task. 
However, there exists a gap between the learned knowledge of PLMs and the goal of CSC task.
PLMs focus on the semantics in text and tend to correct the erroneous characters to semantically proper or commonly used ones, but these aren't the ground-truth corrections.
To address this issue, we propose an \textbf{E}rror-driven \textbf{CO}ntrastive \textbf{P}robability \textbf{O}ptimization (\MethodName{}) framework for CSC task. 
\MethodName{} refines the knowledge representations of PLMs, and guides the model to avoid predicting these common characters through an error-driven way.
Particularly, \MethodName{} is model-agnostic and it can be combined with existing CSC methods to achieve better performance.
Extensive experiments and detailed analyses on SIGHAN datasets demonstrate that \MethodName{} is simple yet effective.

\end{abstract}

\section{Introduction}
\label{Introduction}

\begin{figure}[]
\centering
\includegraphics[width=0.44\textwidth]{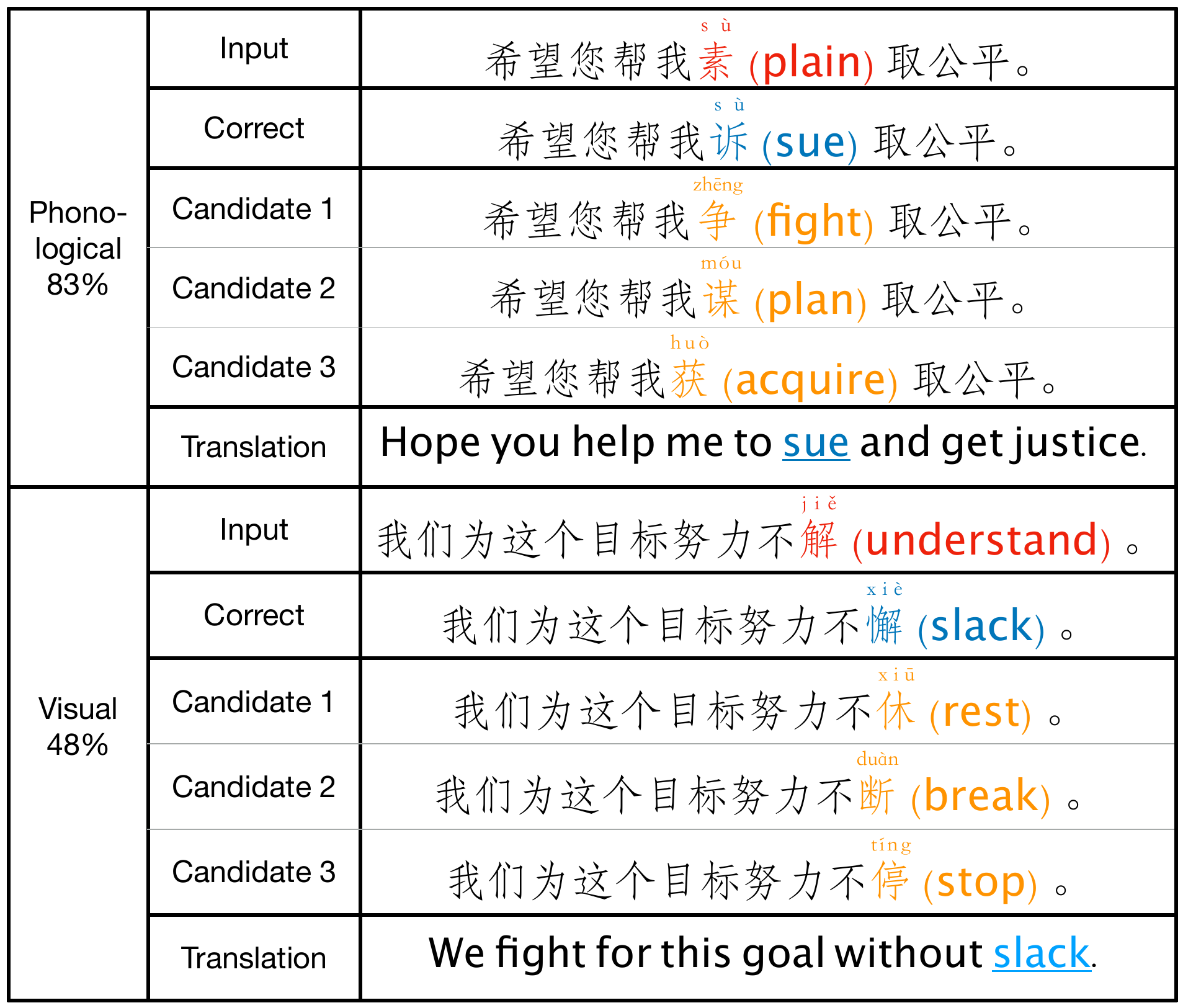}
\caption{Examples of Chinese spelling errors. Previous research~\cite{liu-etal-2021-plome} shows that 83\% of errors belong to phonological error and 48\% belong to visual error. We give the characters with their pronunciation and translation. We mark the \textcolor{red}{input confusing}/\textcolor{blue}{golden}/\textcolor{orange}{common candidate} characters in \textcolor{red}{red}/\textcolor{blue}{blue}/\textcolor{orange}{orange}. The characters in “Candidate” sentences are all predicted by fine-tuned BERT.}
\label{Intro_Figure}
\end{figure}

Chinese Spell Checking (CSC) aims to detect and correct spelling errors in Chinese texts~\cite{wu-etal-2013-integrating}. 
It is a crucial research field for various NLP downstream applications, such as Optical Character Recognition~\cite{afli-etal-2016-using}, search query correction~\cite{gao-etal-2010-large} and essay scoring~\cite{dong-zhang-2016-automatic}. 
However, CSC is also very challenging because it mainly suffers from confusing characters, such as phonologically and visually similar characters~\cite{liu-etal-2010-visually,zhang-etal-2020-spelling}. 
\begin{CJK*}{UTF8}{gbsn}
As illustrated in Figure~\ref{Intro_Figure}, “素 (\pinyin{su4}, plain)” and “诉 (\pinyin{su4}, sue)” are confusing characters for each other due to the shared pronunciation “\pinyin{su4}”.
\end{CJK*}

Recently, pre-trained language models (PLMs) such as BERT~\cite{devlin-etal-2019-bert} have been utilized in the CSC task and became mainstream solutions~\cite{zhang-etal-2020-spelling,cheng-etal-2020-spellgcn}.
However, there exists a significant gap between the learned knowledge of PLMs and the goal of CSC task.
PLMs provide informative representations from the perspective of semantics, but if only considering the semantics in CSC, there are multiple appropriate characters as the correction. 
Without the constraints of phonological and visual similarities, PLMs easily predict semantically proper or common characters due to the masking strategy in the pre-training procedure.

\begin{CJK*}{UTF8}{gbsn}
Figure~\ref{Intro_Figure} presents two predictions of BERT to better understand the gap mentioned before. 
The first example is caused by the misuse of “素 (\pinyin{su4}, plain)” and “诉 (\pinyin{su4}, sue)”.
An ideal CSC model should pay attention to the pronunciation information “\pinyin{su4}” and output the golden character “诉 (sue)” as a correction for the input confusing character.
However, as pre-trained on general corpora, BERT tend to predict semantically proper characters, such as “争 (\pinyin{zheng1}, fight)”, “谋 (\pinyin{mou2}, plan)” and “获 (\pinyin{huo4}, acquire)”.
These characters are also from more commonly used phrases.
In the second example, BERT also overlooks the visual similarity between “解 (\pinyin{jie3}, understand)” and “懈 (\pinyin{xie4}, slack)”, resulting in wrong correction. 
\end{CJK*}

To alleviate this gap, we propose to empower the PLMs to avoid predicting the above-mentioned common characters by optimizing the knowledge representation of PLMs.
Intuitively, if we guide the model to not make the same mistakes it would prone to make before, the model performance should be improved. 
Hence, the mistakes that the model has ever made can be utilized as constraints on the knowledge representation of the model.
In other words, we exploit the past mistakes that the model may make to further enhance the model itself, this is the meaning of our title, “the past mistake is the future wisdom”.

Motivated by the above intuition, we propose the \textbf{E}rror-driven \textbf{CO}ntrastive \textbf{P}robability \textbf{O}ptimization (\MethodName{}), a simple yet effective training framework which aims to refine the knowledge representation of models for CSC. The \MethodName{} consists of two stages: 
(1) \emph{Negative samples selection}. Based on the model's prediction probabilities for different characters, we select the common characters with high probability as negative samples. The golden character is directly regard as the positive sample.
(2) \emph{Contrastive probability optimization}. After obtaining the positive and negative samples, we train the model by Contrastive Probability Optimization (CPO) objective which aims to optimize the prediction probabilities for different characters. Through this optimization process, we can finally narrow the gap between the pre-trained knowledge of PLMs and the goal of CSC. Additionally, \MethodName{} has no strict restrictions on the model to be optimized, so it can further improve the performance of various existing CSC models.

In summary, our contributions are in three folds:
(1) We firstly observe and focus on the negative impact of the gap between the knowledge of PLMs and the goal of CSC.
(2) We propose model-agnostic \MethodName{} framework, which can teach the models to grow and progress with their own past mistakes.
(3) We conduct extensive experiments and detailed analyses on SIGHAN benchmarks and achieve state-of-the-art performance.

\begin{figure*}[]
\centering
\includegraphics[height=0.48\textwidth]{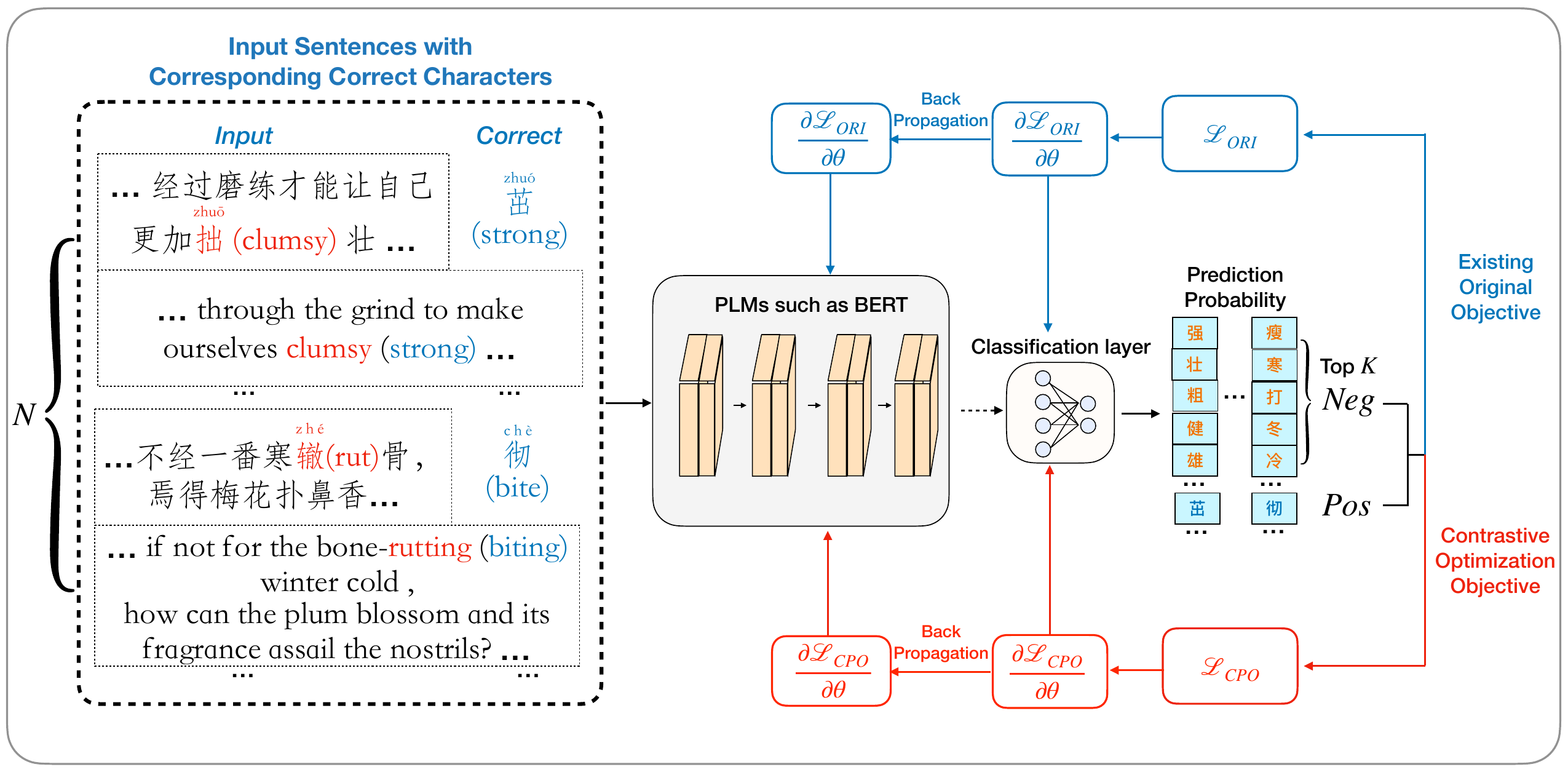}
\caption{\begin{CJK*}{UTF8}{gbsn}Overview of \MethodName{} framework. We select negative samples according to the original prediction probability of PLMs (e.g, for the position of “拙”, PLMs predicts the Top 5 characters as “强”, “壮”, “粗”, “健”, and “雄”.), then optimize the PLMs with the contrastive optimization objective and traditional original objective.\end{CJK*}}
\label{Method_Figure}
\end{figure*}

\section{Related Work}
\subsection{Chinese Spell Checking}
Early works in CSC mainly focus on designing heuristic rules to detect different kinds of errors~\cite{chang-etal-2015-introduction,chu-lin-2015-ntou}.
Most of these methods rely on solid linguistic knowledge and manually designed features, and thus do not have the generalization performance required for large-scale application. 
Next, various traditional machine learning algorithms, such as Conditional Random Field (CRF) and Hidden Markov Model (HMM), are applied in CSC~\cite{wang-liao-2015-word, zhang-etal-2015-hanspeller}. Then, deep learning-based models have gradually become the mainstream of CSC in recent years~\cite{wang-etal-2021-dynamic, guo-etal-2021-global, zhang-etal-2021-correcting}.

\citet{wang-etal-2018-hybrid} utilize a BiLSTM trained on an automatically generated dataset to convert CSC to sequence labeling problem. 
\citet{hong-etal-2019-faspell} propose to generate and curtail the candidate characters through a BERT-based denoising autoencoder.
The Soft-Masked BERT model~\cite{zhang-etal-2020-spelling} uses two separate networks for detection and correction.
Then SpellGCN~\cite{cheng-etal-2020-spellgcn} uses GCN~\cite{kipf2017semi} to fuse character embedding with similar pronunciation and shape, explicitly modeling the relationship between characters.
PLOME~\cite{liu-etal-2021-plome} is proposed to be a task-specific pre-trained language model for CSC, which designs a confusion set based masking strategy and introduces various external knowledge.
Additionally, REALISE~\cite{xu-etal-2021-read} verifies that the multimodal knowledge can be leveraged to improve CSC performance.

\subsection{Contrastive Learning}
The main motivation of contrastive learning is to attract the positive samples and repulse the negative samples in a certain space~\cite{HadsellCon, chen2020simple, khosla2020supervised}. Existing contrastive learning models in NLP are mainly focusing on the \emph{language representation space} (e.g, word/sentence/semantic representations)~\cite{iter-etal-2020-pretraining,gao2021simcse,wang-etal-2021-cline}. Different from them, our proposed method directly optimizes the model's \emph{probability space} for different characters through selected positive/negative samples and their original predicted probability. 

\section{Methodology}
\label{Methodology}
In this section, we introduce the proposed \MethodName{} in details, as illustrated in Figure~\ref{Method_Figure}. 
\MethodName{} aims to refine the knowledge representation of PLMs to narrow the gap between it and the essential of CSC task.
As mentioned in Section~\ref{Introduction}, with the model before our optimization process, we select the mistakes generated by this model itself to be the negative samples. Then through the Contrastive Probability Optimization objective, we maximize the prediction probabilities of the model for correct answers and minimize the prediction probabilities of the model for negative samples. In this error-driven way, the original prediction probabilities of the model are refined, improving the performance of the model on the CSC task. \emph{Therefore, the model will grow and progress after making mistakes again and again, just as humans do.}
Note that the proposed \MethodName{} is a model-agnostic framework, we can choose different PLMs or CSC models to be optimized in practice for better performance.

\subsection{Observation and Intuition}
\label{Observation}
To present our approach more clearly, we first describe our observation, and then give our explanation of the observation and intuition.

The key observation that \MethodName{} builds on is that PLMs such as BERT cannot focus well on the confusing characters that need to be paid more attention in the CSC task, as illustrated in Figure~\ref{Intro_Figure}.
We think that this gap comes mainly from the general corpora and the training paradigm used in the pre-training of language models. Taking the BERT as an example, its pre-training corpus is mainly from the text in Wikipedia, which has a very low proportion of contexts containing confusing characters, as verfied in Section~\ref{Case_Study}. Additionally,~\citet{devlin-etal-2019-bert} randomly choose 15\% of tokens in the entire corpus to be masked by a fixed token “[MASK]” and then recover them. This masking-recovering strategy makes the knowledge acquired by PLMs in pre-training process discontinuous in the CSC task~\cite{liu-etal-2021-plome}. Because the size of confusing characters will be lower in the 15\% of characters that are randomly selected.

In fact, there also exists the same challenge when humans correct spelling errors. When only given the context of input sentence without seeing the misspelling, they tend to associate the common character rather than the confusing character with the context.
Therefore, humans or models would wrongly predict common characters. \emph{Intuitively, if the model can be optimized with common characters through an error-driven way, then the model can certainly be further enhanced, just as humans get progress from the mistakes they have made.}

\subsection{Stage 1: Negative Samples Selection} 
We define the negative samples in CSC as those common characters that be incorrectly assigned high prediction probability by PLMs before our optimization process. 
According to our observation, negative samples that can form common collocations or phrases with the context tend to be assigned higher probability than the golden character, leading the model to make wrong corrections.
Therefore, we use a simple strategy based on the prediction probability to select the negative samples which we utilize in the next stage.

Specifically, we use PLMs such as BERT to predict the original character for each input token based on the output of the last transformer layer. The prediction probability of the i-th token $x_i$ in a sentence $X$ is defined as:
\begin{equation}
    p\left(y_i = j \mid X\right)=\operatorname{softmax}\left(\boldsymbol{W} \boldsymbol{h}_{\boldsymbol{i}} + \boldsymbol{b}\right)[j],
\end{equation}
where $p\left( y_i = j \mid X \right)$ means the conditional probability that the i-th token $x_i$ is predicted as the j-th character in the vocabulary of PLMs, $\boldsymbol{W} \in \boldsymbol{R}^{vocab \times hidden}$ and $\boldsymbol{b} \in \boldsymbol{R}^{vocab}$ are learnable parameters, $vocab$ is the size of vocabulary and the $hidden$ is the size of hidden state, $\boldsymbol{h_i} \in \boldsymbol{R}^{hidden}$ is hidden state output of PLMs for the i-th token $x_i$.

Based on the original prediction probability, if the model makes wrong correction for the input character, we will select negative samples for the input character. 
The negative samples set $Neg$ is selected from the candidate set $T$ as:
\begin{gather}
    T = \{ t \mid t\in V \ \ and \ \  t \neq t^+ \}, \\
    Neg = \argmax_{T^{\prime} \subset T, \left| T^{\prime} \right| = K} \sum_{t^- \in T^{\prime}} p\left(y_i = t^- \mid X\right),
   \label{Negative_Samples}
\end{gather}
where $t^-$ and $t^+$ mean the negative and positive samples, respectively.
The negative samples $t^-$ are selected from those tokens whose prediction probability is in the Top $K$ of the vocabulary $V$, and the best value of $K$ is selected empirically. It is worth noting that the training process is supervised in the CSC task, so we can regard the golden character as the positive sample $t^+$.

\subsection{Stage 2: Contrastive Probability Optimization} 
\label{CPO_Method}
 After obtaining the positive/negative samples and their corresponding prediction probability, we train the model by Contrastive Probability Optimization (CPO) objective which is defined as:
\begin{equation}
\begin{aligned}
\mathcal{L}_{CPO}= - \frac{1}{N} \sum_{i=1}^{N} \frac{1}{K} \sum_{k=1}^{K} \{ p\left(y_i = t^+ \mid X \right) \\ - p\left(y_i = t_k^- \mid X \right) \},
\end{aligned}
\end{equation}
where $N$ is the batch size, $K$ is the selected negative samples size, $t_k^-$ is the $k$-th negative sample in $Neg$. The CPO objective aims to teach the model to increase the prediction probability for positive sample (i.e., confusing character) and decrease the prediction probabilities for negative samples (i.e., common characters)  by the maximum likelihood of the difference between the original probabilities for positive and negative samples.

To preserve the generalization performance of the model, we train both the existing original objective $\mathcal{L}_{ORI}$ and the CPO objective $\mathcal{L}_{CPO}$. The overall objective is defined as:

\begin{equation}
    \mathcal{L} = \lambda_{1} \mathcal{L}_{ORI} + \lambda_{2} \mathcal{L}_{CPO},
\label{joint_objective}
\end{equation}
where $\lambda_{1}$ and $\lambda_{2}$ are weighting factors for two objectives. We use cross-entropy loss function as the $\mathcal{L}_{ORI}$ for BERT in our experiments. The training pseudo-code of \MethodName{} is shown in Appendix~\ref{Appendix_A}. As described in Equation~\ref{joint_objective}, we can replace the $\mathcal{L}_{ORI}$ with other models' training objectives, so \MethodName{} is model-agnostic and it can be easily used in other PLMs or previous CSC methods to achieve further improvements.

Most previous works use $\operatorname{softmax}$ and $\operatorname{cross-entropy}$ functions to train CSC models. But why just using $\operatorname{softmax}$ is not enough and using CPO is necessary?
\textbf{Theoretically}: 
(1) \emph{Their motivations are different}, $\operatorname{softmax}$ is to normalize the PLMs' logits into a probability distribution, but CPO aims to refine the knowledge representation of PLMs in the probability space. 
(2) \emph{Their scopes are different}, $\operatorname{softmax}$ relies on all logits output by models for weighted calculation, this global weighting mechanism makes it not have good local attention. However, CPO can pay attention to a part of really difficult samples that models would often make mistakes through the negative samples selection stage. 
(3) \emph{Their results are different}, through the $\operatorname{softmax}$ operation, we finally obtain a probability distribution that is softer than the original logits. But the CPO we proposed can eventually change the order of the original prediction probability, directing the model to assign higher probability to positive sample and lower probabilities to negative samples. Therefore, our work can be regarded as a great complement to the traditional $\operatorname{softmax}+\operatorname{cross-entropy}$ training paradigm.
\textbf{Empirically}, we conducted in-depth analyses in Sections~\ref{statistics_exp}-~\ref{lambda_exp}.

\section{Experiments}
\label{Exper}
In this section, we introduce the details of experiments and main results we obtained. Then we conduct detailed analyses and discussions to verify the effectiveness of our method.

\subsection{Datasets}
\paragraph{Training Data} We use the same training data by following previous works~\cite{zhang-etal-2020-spelling,liu-etal-2021-plome,xu-etal-2021-read}, including the training samples from SIGHAN13~\cite{wu-etal-2013-chinese}, SIGHAN14~\cite{yu-etal-2014-overview}, SIGHAN15~\cite{tseng-etal-2015-introduction} and the pseudo training samples (size of 271K, we denote this part of samples as Wang271K in our paper) automatically generated by OCR-based and ASR-based methods~\cite{wang-etal-2018-hybrid}. 

\paragraph{Test Data} To ensure the fairness, we use the exact same test data as the baseline methods, from the test datasets of SIGHAN13/14/15. Noted that the text of original SIGHAN datasets is in the Traditional Chinese, we pre-process these original datasets to the Simplified Chinese using the OpenCC\footnote{https://github.com/BYVoid/OpenCC}. This data conversion procedure has been widely used in previous works~\cite{wang-etal-2019-confusionset,cheng-etal-2020-spellgcn,zhang-etal-2020-spelling}. The detailed statistic of the training/test data we use in our experiments is presented in Appendix~\ref{Appendix_B}.

\subsection{Baseline Methods}
To evaluate the performance of \MethodName{}, we select several advanced strong baseline methods: 
\textbf{BERT}~\cite{devlin-etal-2019-bert} is directly fine-tuned on the training data.
\textbf{Hybrid}~\cite{wang-etal-2018-hybrid} casts CSC into sequence labeling problem and implements BiLSTM model. 
\textbf{FASpell}~\cite{hong-etal-2019-faspell} consists of a denoising autoencoder and a decoder. 
\textbf{Soft-Masked BERT}~\cite{zhang-etal-2020-spelling} consists of a detection network and a correction network. 
\textbf{SpellGCN}~\cite{cheng-etal-2020-spellgcn} integrates the confusion set to the correction model through GCNs.
\textbf{PLOME}~\cite{liu-etal-2021-plome} is a task-specific PLM which jointly learns how to understand language and correct spelling errors.
\textbf{REALISE}~\cite{xu-etal-2021-read} is a multimodel model which captures and mixes the semantic, phonetic and graphic information to improve CSC performance. REALISE is the previous state-of-the-art method on SIGHAN13/14/15 datasets.

\begin{table*}[h]
\small
\centering
\begin{tabular}{@{}c|l|p{0.65cm}p{0.65cm}p{0.65cm}p{0.65cm}|p{0.65cm}p{0.65cm}p{0.65cm}p{0.65cm}@{}}
\toprule
\multirow{2}{*}{Dataset} & \multicolumn{1}{c|}{\multirow{2}{*}{Method}} & \multicolumn{4}{c|}{Detection Level} & \multicolumn{4}{c}{Correction Level} \\
 & \multicolumn{1}{c|}{} & Acc & Pre & Rec & F1 & Acc & Pre & Rec & F1 \\ \midrule
 
\multicolumn{1}{l|}{\multirow{9}{*}{SIGHAN13}} & Hybrid~\cite{wang-etal-2018-hybrid} & - & 54.0 & 69.3 & 60.7 & - & - & - & 52.1 \\
\multicolumn{1}{l|}{} & FASpell~\cite{hong-etal-2019-faspell} & 63.1 & 76.2 & 63.2 & 69.1 & 60.5 & 73.1 & 60.5 & 66.2 \\
\multicolumn{1}{l|}{} & SpellGCN~\cite{cheng-etal-2020-spellgcn} & - & 80.1 & 74.4 & 77.2 & - & 78.3 & 72.7 & 75.4 \\
\cmidrule(l){2-10} 
\multicolumn{1}{l|}{} & BERT~\cite{xu-etal-2021-read} & 77.0 & 85.0 & 77.0 & 80.8 & 77.4 & 83.0 & 75.2 & 78.9 \\
\multicolumn{1}{l|}{} & \MethodName{}~(BERT) & 81.7$^\uparrow$ & 87.2$^\uparrow$ & 81.7$^\uparrow$ & 84.4$^\uparrow$ & 80.7$^\uparrow$ & 86.1$^\uparrow$ & 80.6$^\uparrow$ & 83.3$^\uparrow$ \\ 
\cmidrule(l){2-10} 
\multicolumn{1}{l|}{} & REALISE~\cite{xu-etal-2021-read} & \underline{82.7} & \underline{88.6} & \underline{82.5} & \underline{85.4} & \underline{81.4} & \underline{87.2} & \underline{81.2} & \underline{84.1} \\
\multicolumn{1}{l|}{} & \MethodName{}~(REALISE) & \textbf{83.3}$^\uparrow$ & \textbf{89.3}$^\uparrow$ & \textbf{83.2}$^\uparrow$ & \textbf{86.2}$^\uparrow$ & \textbf{82.1}$^\uparrow$ & \textbf{88.5}$^\uparrow$ & \textbf{82.0}$^\uparrow$ & \textbf{85.1}$^\uparrow$ \\

\midrule

\multicolumn{1}{l|}{\multirow{9}{*}{SIGHAN14}} & Hybrid~\cite{wang-etal-2018-hybrid} & - & 51.9 & 66.2 & 58.2 & - & - & - & 56.1 \\
\multicolumn{1}{l|}{} & FASpell~\cite{hong-etal-2019-faspell} & 70.0 & 61.0 & 53.5 & 57.0 & 69.3 & 59.4 & 52.0 & 55.4 \\
\multicolumn{1}{l|}{} & SpellGCN~\cite{cheng-etal-2020-spellgcn} & - & 65.1 & 69.5 & 67.2 & - & 63.1 & 67.2 & 65.3 \\ 
\cmidrule(l){2-10} 
\multicolumn{1}{l|}{} & BERT~\cite{xu-etal-2021-read} & 75.7 & 64.5 & 68.6 & 66.5 & 74.6 & 62.4 & 66.3 & 64.3 \\
\multicolumn{1}{l|}{} & \MethodName{}~(BERT) & 76.7$^\uparrow$ & 65.8$^\uparrow$ & 69.0$^\uparrow$ & 67.4$^\uparrow$ & 75.7$^\uparrow$ & 63.7$^\uparrow$ & 66.9$^\uparrow$ & 65.3$^\uparrow$ \\ 
\cmidrule(l){2-10} 
\multicolumn{1}{l|}{} & REALISE~\cite{xu-etal-2021-read} & \underline{78.4} & \underline{67.8} & \underline{71.5} & \underline{69.6} & \underline{77.7} & \underline{66.3} & \underline{70.0} & \underline{68.1} \\
\multicolumn{1}{l|}{} & \MethodName{}~(REALISE) & \textbf{79.0}$^\uparrow$ & \textbf{68.8}$^\uparrow$ & \textbf{72.1}$^\uparrow$ & \textbf{70.4}$^\uparrow$ & \textbf{78.5}$^\uparrow$ & \textbf{67.5}$^\uparrow$ & \textbf{71.0}$^\uparrow$ & \textbf{69.2}$^\uparrow$ \\ 
\midrule

\multicolumn{1}{l|}{\multirow{12}{*}{SIGHAN15}} & Hybrid~\cite{wang-etal-2018-hybrid} & - & 56.6 & 69.4 & 62.3 & - & - & - & 57.1 \\
\multicolumn{1}{l|}{} & FASpell~\cite{hong-etal-2019-faspell} & 74.2 & 67.6 & 60.0 & 63.5 & 73.7 & 66.6 & 59.1 & 62.6 \\
\multicolumn{1}{l|}{} & SpellGCN~\cite{cheng-etal-2020-spellgcn} & - & 74.8 & 80.7 & 77.7 & - & 72.1 & 77.7 & 75.9 \\ 
\multicolumn{1}{l|}{} & PLOME~\cite{liu-etal-2021-plome} & - & \underline{77.4} & \underline{81.5} & \underline{79.4} & - & 75.3 & 79.3 & 77.2 \\
\cmidrule(l){2-10} 
\multicolumn{1}{l|}{} & Soft-Masked BERT~\cite{zhang-etal-2020-spelling} & 80.9 & 73.7 & 73.2 & 73.5 & 77.4 & 66.7 & 66.2 & 66.4 \\
\multicolumn{1}{l|}{} & \MethodName{}~(Soft-Masked BERT) & 81.2$^\uparrow$ & 74.0$^\uparrow$ & 76.6$^\uparrow$ & 75.3$^\uparrow$ & 79.1$^\uparrow$ & 67.0$^\uparrow$ & 72.3$^\uparrow$ & 69.6$^\uparrow$ \\
\cmidrule(l){2-10} 
\multicolumn{1}{l|}{} & BERT~\cite{xu-etal-2021-read} & 82.4 & 74.2 & 78.0 & 76.1 & 81.0 & 71.6 & 75.3 & 73.4 \\
\multicolumn{1}{l|}{} & \MethodName{}~(BERT) & \textbf{85.5}$^\uparrow$ & \textbf{78.2}$^\uparrow$ & 82.3$^\uparrow$ & \textbf{80.2}$^\uparrow$ & \textbf{84.6}$^\uparrow$ & \textbf{76.6}$^\uparrow$ & 80.4$^\uparrow$ & 78.4$^\uparrow$ \\
\cmidrule(l){2-10} 
\multicolumn{1}{l|}{} & REALISE~\cite{xu-etal-2021-read} & \underline{84.7} & 77.3 & 81.3 & 79.3 & \underline{84.0} & \underline{75.9} & \underline{79.9} & \underline{77.8} \\
\multicolumn{1}{l|}{} & \MethodName{}~(REALISE) & 85.0$^\uparrow$ & 77.5$^\uparrow$ & \textbf{82.6}$^\uparrow$ &  80.0$^\uparrow$ & 84.2$^\uparrow$ & 76.1$^\uparrow$ & \textbf{81.2}$^\uparrow$ & \textbf{78.5}$^\uparrow$ \\

\bottomrule
\end{tabular}

\caption{The performance of \MethodName{} and all baseline methods. Note that all baseline results are directly from other published paper. \MethodName{}~(model-X) means that we perform \MethodName{} framework on model-X. We underline the previous state-of-the-art performance for convenient comparison. “$^\uparrow$” indicates that the corresponding baseline method receives a further performance improvement after optimization by \MethodName{}. }
\label{Main_Results}
\end{table*}

\subsection{Experimental Setup}
In terms of evaluation granularity, there are two levels of metrics, namely character/sentence-level. Obviously, the sentence-level metric is stricter than the character-level metric because there may be multiple wrong characters in a sentence. One sentence sample is considered to be correct only when all the wrong characters in it are detected and corrected successfully. Therefore, we report the sentence-level metrics for evaluation, which are widely used in previous works~\cite{li-etal-2021-exploration,huang-etal-2021-phmospell,xu-etal-2021-read}. 

Specifically, the metrics we report include Accuracy, Precision, Recall and F1 score for detection and correction levels. 
At the detection level,  all locations of wrong characters in a sentence should be identical successfully. 
At the correction level, the model must not only detect but also correct all the erroneous characters with the gold standard.

Other implementation details and hyper-parameters choices are presented in Appendix~\ref{Appendix_D}.

\begin{figure*}[]
\centering
\includegraphics[height=0.37\textwidth]{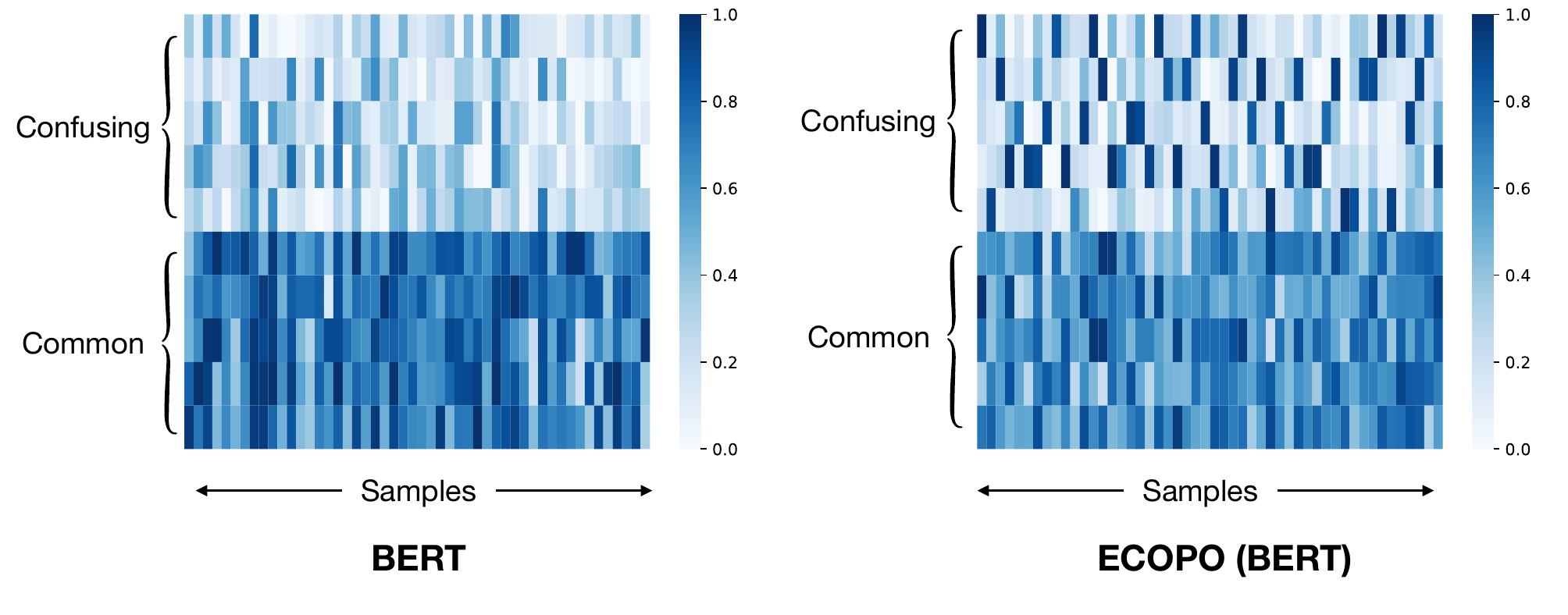}
\caption{Heat map visualization of probability. The darker the blue, the higher the model's prediction probability for a particular character (vertical axis) given the input of samples containing misspelled characters (horizontal axis). The selected samples are from SIGHAN15, and the original BERT would make wrong corrections for them.}
\label{Heat_Map}
\end{figure*}

\subsection{Experimental Results}
From Table~\ref{Main_Results}, we can observe that:
\begin{enumerate}
    \item The \MethodName{}~(BERT) performs better than BERT on all test sets and evaluation metrics. Specifically, \MethodName{}~(BERT) achieves significant improvement on SIGHAN15, and outperforms the previous state-of-the-art models with a very thin model, while REALISE and PLOME are two complex models with some auxiliary modules. Note that \MethodName{}~(BERT) only consists of a BERT encoder.
    \item From the results on the SIGHAN14 test set, we can see that the performance improvement of \MethodName{}~(BERT) based on BERT is not as large as on the other two test sets, but still effective. Additionally, due to the model-agnostic advantage of \MethodName{}, it can be simply combined with other previous state-of-the-art models such as REALISE and get further enhancement, which are presented in the rows of REALISE and \MethodName{}~(REALISE).
    \item Considering the impact of external knowledge, several previous works exploit various additional information to improve performance. For example, FASpell and SpellGCN introduce character similarity to CSC, REALISE and PLOME propose to leverage multimodal knowledge such as phonetic and graphic information. Unlike the aforementioned models, \MethodName{}~(BERT) achieves competitive performance without any additional knowledge and optimizing only based on the mistakes that the original BERT itself has made.
    \item To verify the model-agnostic characteristic of \MethodName{}, we choose two other models including Soft-Masked BERT and REALISE to be optimized. Practically, we train the combined model with the joint objective, as described in Equation~\ref{joint_objective}. From the results of Table~\ref{Main_Results}, we can see that \MethodName{}'s improvement is stable and significant over the three models. 
\end{enumerate}

\subsection{Analysis and Discussion}

\subsubsection{Statistics of Different Characters}
\label{statistics_exp}
To further empirically explain why the method we proposed is effective, we conduct sufficient statistical experiments, as shown in Table~\ref{Statistics_Results}. 
We apply different methods to the SIGHAN13/14/15 datasets, and carry out statistical analyses on their wrong correction samples. Note that if a character co-occurs with the character before or after the error position more than 1,000 times in wiki2019zh\footnote{The general pre-training corpus which is from Wikipedia dump (as of February 7, 2019) and contains one million pages. }, we regard it as a common character. 

From Table~\ref{Statistics_Results}, we can see that when only $\operatorname{softmax}$ is used, most of the failures of the model are because it incorrectly assigns higher prediction probabilities to common characters, which reflects the gap between the pre-trained knowledge of PLMs and the goal of CSC. When we run \MethodName{} or only CPO, the model does pay more attention to the less common but more confusing characters. Our proposed CPO indeed effectively change the model's predictions for different types of characters. Thus, CPO refines the knowledge representation of PLMs for CSC and narrow the gap between PLMs and CSC, but $\operatorname{softmax}$ does not.

\begin{table}[h]
\small
\centering
\begin{tabular}{@{}c|c|c|c@{}}
\toprule
Dataset & Method & Common & Confusing \\
\midrule
 
\multicolumn{1}{l|}{\multirow{4}{*}{SIGHAN13}} & $\operatorname{softmax}$ & 172 (76\%) & 54 (24\%) \\
\cmidrule(l){2-4} 
\multicolumn{1}{l|}{} & CPO   & 108 (54\%) & 92 (46\%) \\
\cmidrule(l){2-4} 
\multicolumn{1}{l|}{} & \MethodName{} & 100 (52\%) & 93 (48\%) \\
\midrule

\multicolumn{1}{l|}{\multirow{4}{*}{SIGHAN14}} & $\operatorname{softmax}$  & 208 (77\%) & 62 (23\%) \\
\cmidrule(l){2-4} 
\multicolumn{1}{l|}{}  & CPO   & 159 (61\%) & 101 (39\%) \\
\cmidrule(l){2-4} 
\multicolumn{1}{l|}{}  & \MethodName{} & 152 (59\%) & 106 (41\%) \\
\midrule

\multicolumn{1}{l|}{\multirow{4}{*}{SIGHAN15}}  & $\operatorname{softmax}$  & 171 (82\%) & 38 (18\%) \\
\cmidrule(l){2-4} 
\multicolumn{1}{l|}{}  & CPO & 72 (41\%) & 103 (59\%) \\
\cmidrule(l){2-4} 
\multicolumn{1}{l|}{}  & \MethodName{} & 68 (40\%)  & 101 (60\%) \\

\bottomrule
\end{tabular}

\caption{Statistical results on different types of characters. The statistical samples are the all wrong correction samples of different methods.}
\label{Statistics_Results}
\end{table}

\subsubsection{Visualization of Common/Confusing Character Probability}
\label{visualization_exp}

The key objective of \MethodName{} is to optimize the prediction probability of the PLMs for two different kinds of characters, i.e., \textbf{common characters} which original PLMs would be more inclined and \textbf{confusing characters} which CSC task should pay more attention to.
Therefore, we visualize the probability optimization effect of \MethodName{} in this part of experiment. 
Specifically, we apply BERT and \MethodName{}~(BERT) to predict the character which should appear at the position of the misspelled character based on its context. 
We select the Top-5 characters co-occurring with the context of the misspelled character as the common characters, and 5 confusing characters from the widely used confusion set~\cite{wu-etal-2013-chinese}. Note that we ensure that the common and confusing characters selected are not duplicated, and the golden character must be in the selected 5 confusing characters. Then we visualize the prediction probabilities of common/confusing characters as a heat map. 

Figure~\ref{Heat_Map} shows the prediction probability distributions of BERT and \MethodName{}~(BERT) for the common/confusing characters. 
By comparison, we can see that BERT assigns higher probability to common characters than confusing characters, and \MethodName{}~(BERT) focuses more on confusing characters which are similar to the golden character.
This difference in BERT before and after \MethodName{}'s optimization is consistent with our study motivation and design objective. We can see that \MethodName{} does refine the knowledge representation and prediction probability of BERT for different characters. 

\subsubsection{Effects of Weighting Factors $\lambda_{1}$,$\lambda_{2}$}
\label{lambda_exp}

\begin{figure} 
\centering
\subfigure[Detection performance] { \label{lambda_detection} 
\includegraphics[height = 0.40 \columnwidth,width=0.46\columnwidth]{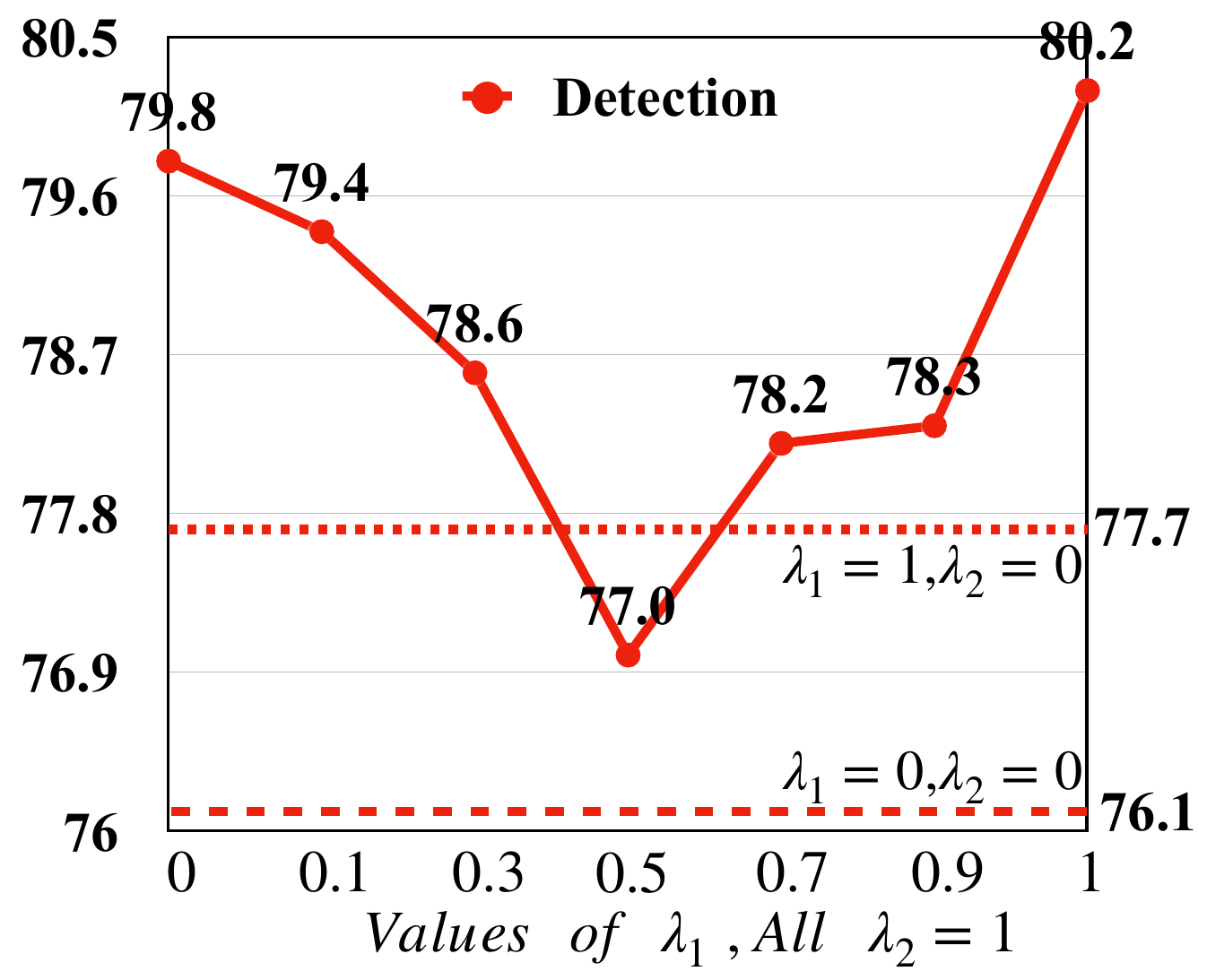} 
} 
\subfigure[Correction performance] { \label{lambda_correction} 
\includegraphics[height = 0.40 \columnwidth, width=0.46\columnwidth]{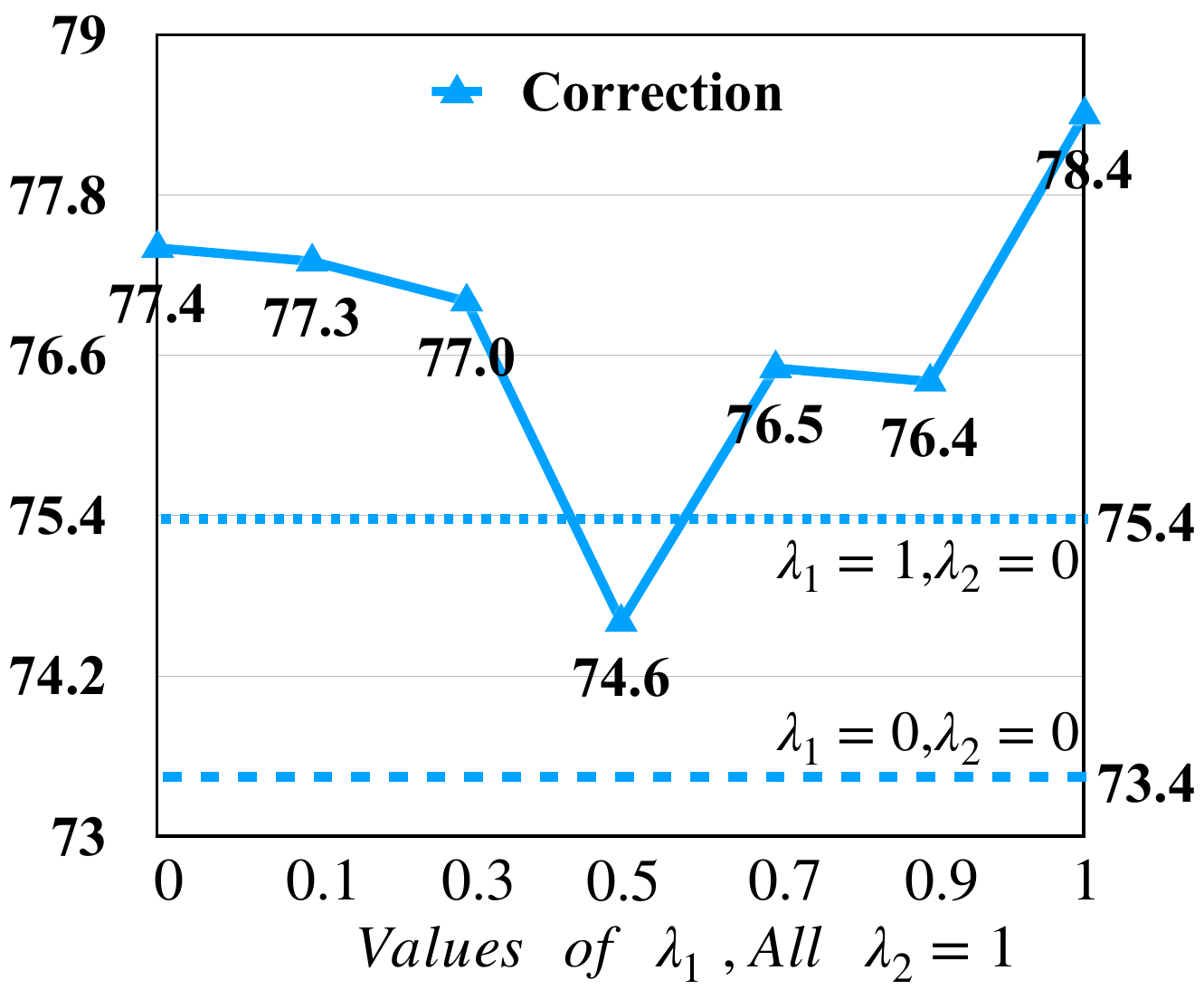} 
} 
\caption{The F1 results on SIGHAN15, using different combinations of $\lambda_{1}$,$\lambda_{2}$ in Equation~\ref{joint_objective} in \MethodName{}~(BERT). 
When $\lambda_{1}=0$,$\lambda_{2}=0$, it is equivalent to the baseline BERT. } 
\label{lambda_results} 
\end{figure}

Firstly, from Figure~\ref{lambda_results}, we can see that no matter how the values of $\lambda_{1}$,$\lambda_{2}$ change, \MethodName{}~(BERT) always has improvement compared to the baseline BERT, which reflects the general effectiveness of our proposed method. We also can find that whether only using $\mathcal{L}_{ORI}$ ($\lambda_{1}=1$,$\lambda_{2}=0$) or $\mathcal{L}_{CPO}$ ($\lambda_{1}=0$,$\lambda_{2}=1$) for training, there is an improvement compared to the baseline model. Besides, only using $\mathcal{L}_{CPO}$  has a greater improvement than only using $\mathcal{L}_{ORI}$, which illustrates the advantage of our proposed CPO over $\operatorname{softmax}$. Furthermore, when $\lambda_{2}$ is fixed to 1, as $\lambda_{1}$ increases, the model performance shows a trend of first decreasing and then increasing. 
From this phenomenon, we suspect that the widely used $\mathcal{L}_{ORI}$ in previous works has a certain regularization effect on the probability space of the model. Also for this reason, only using $\mathcal{L}_{ORI}$ has improvements compared to the baseline. Additionally, the regularization effect of $\mathcal{L}_{ORI}$ is good for the process of $\mathcal{L}_{CPO}$ optimizing the probability representation, and can help model avoid over-fitting.
Therefore, in practice, we chose the combination that perform best in SIGHAN13/14/15, namely $\lambda_{1}=1$,$\lambda_{2}=1$.

\subsubsection{Effects of Negative Samples Size $K$} 
\label{K_exp}
\begin{figure} 
\centering

\includegraphics[width=0.88\columnwidth]{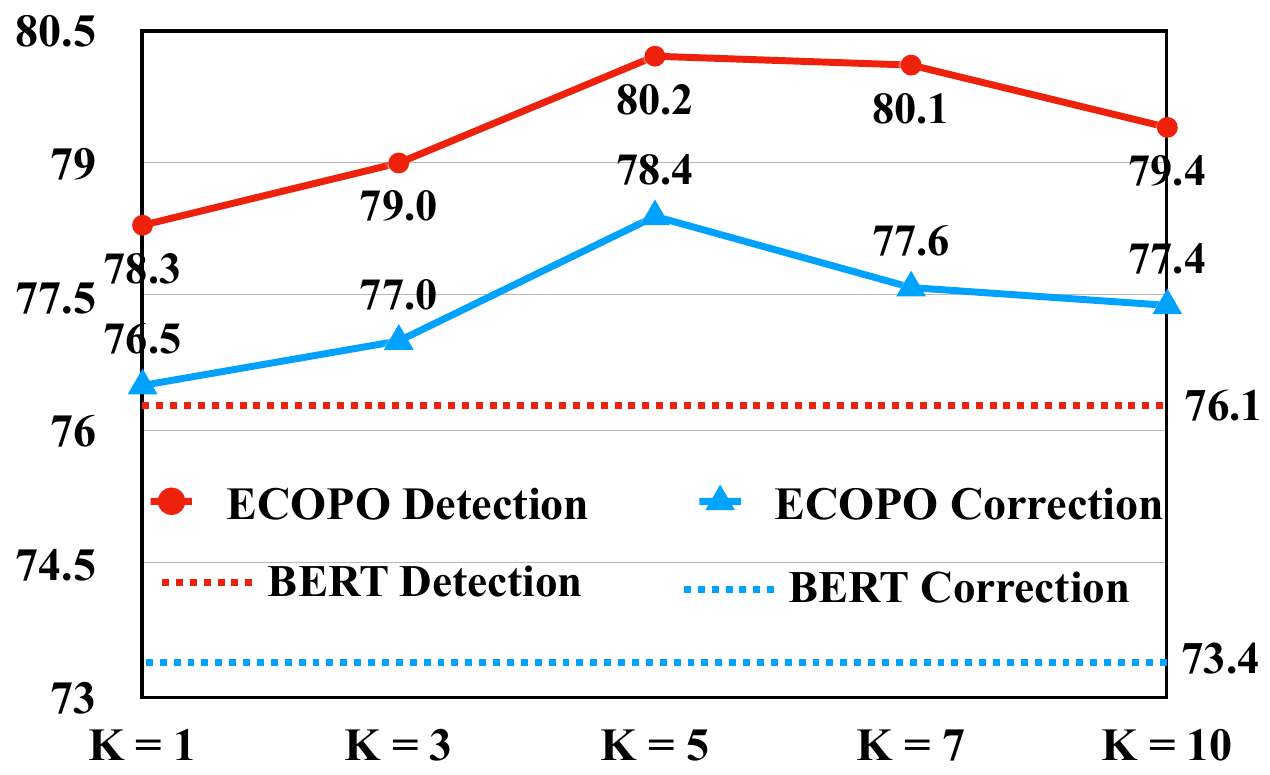} 

\caption{The F1 results on SIGHAN15, using different values of $K$ in Equation~\ref{Negative_Samples} in \MethodName{}~(BERT). 
The dotted lines represent the baseline BERT's performance.}
\label{Different_K_Results} 
\end{figure} 

As different amounts of negative samples can affect \MethodName{}'s performance, it is essential to study the impact of negative samples size $K$ in Equation~\ref{Negative_Samples}.

Figure~\ref{Different_K_Results} illustrates the performance change from the perspective of detection and correction. 
We find that when the value of $K$ reaches a certain value (e.g., $ K > 5$), the overall performance of the model (F1 score) does not improve anymore. This is because \MethodName{} optimizes the model based on the probability representation, when the value of $K$ becomes very large, the predicted probabilities of samples become so small that they have almost no effect on the probability optimization of the positive sample. Therefore, choosing an appropriate $K$ value is critical to the performance improvement of \MethodName{}, although \MethodName{} has significant improvement based on BERT at all values of $K$.

\subsection{Case Study for Probability Optimization}
\label{Case_Study}
\begin{CJK*}{UTF8}{gbsn}

\begin{table}[t]
\small
\centering
\begin{tabular}{p{1.6cm}p{4.8cm}}
\toprule

% Case 1: \\ \midrule
\textbf{Input:} & 与其自暴自\textcolor{red}{气} (\textcolor{blue}{弃}) 不如往好处想。 \\
& It's better to think for the good than to \textcolor{red}{be angry} (\textcolor{blue}{give up}). \\ 
\midrule
\textbf{BERT:} & [\textcolor{orange}{己 (own)}, \textcolor{orange}{大 (big)}, \textcolor{orange}{利 (benefit)}] \\
\textbf{\MethodName{}:} & [\textcolor{blue}{弃 (give up)}, \textcolor{orange}{尊 (respect)}, \textcolor{orange}{强 (strong)}] \\ \midrule \midrule
  
% Case 2: \\ \midrule
\textbf{Input:} & 我努力打败数不\textcolor{red}{进} (\textcolor{blue}{尽})的风雨。 \\
& I try to beat the \textcolor{red}{enter} (\textcolor{blue}{endless}) storms. \\ 
\midrule
\textbf{BERT:} & [\textcolor{orange}{起 (raise)}, \textcolor{orange}{上 (up)}, \textcolor{orange}{得 (get)}]  \\
\textbf{\MethodName{}:} & [\textcolor{blue}{尽 (endless)}, \textcolor{orange}{得 (get)}, \textcolor{orange}{完 (end)}] \\ \bottomrule

\end{tabular}
\caption{Examples of spelling errors and corresponding output (Top 3 candidates) of original BERT and \MethodName{}~(BERT). 
We mark the \textcolor{red}{input confusing}/\textcolor{blue}{golden}/\textcolor{orange}{wrong correction} characters in \textcolor{red}{red}/\textcolor{blue}{blue}/\textcolor{orange}{orange}.  }
\label{Case_Studies}
\end{table}

Table~\ref{Case_Studies} shows the comparisons between the correction results of BERT and \MethodName{}~(BERT).
In the first examples, the output of BERT such as “己”, “大” and “利” all can form a correct Chinese phrase with “自”, but they cause a semantic incoherence for the whole sentence. 
The statistics of the general pre-training corpus wiki2019zh show that “自己” co-occurs 136,318 times and “自弃” co-occurs 119 times, which verifies the intuition about common/confusing characters described in Section~\ref{Observation}.
In the second example as well, the output of BERT can be formed with “数不” as reasonable phrases.
From the two examples, we can see that \MethodName{} does guide the BERT to accurately predict the ideal confusing characters by the highest probability and make the right corrections. Such experimental results are in line with our work's core motivation.

\end{CJK*}

\section{Conclusion}
In this paper, we introduce to promote the CSC task by narrowing the gap between the knowledge of PLMs and the goal of CSC. We propose the \MethodName{}, a simple yet effective training framework that aims to perform an error-driven optimization for the PLMs based on their original probability representation. Extensive experiments and empirical results show the competitive performance of our method. In the future, we will study how to automatically measure the quality of negative samples to further enhance our method. Additionally, applying our core idea and motivation to kinds of other tasks will be an interesting direction.

\section{Acknowledgement}
This research is supported by National Natural Science Foundation of China (Grant No. 6201101015), Beijing Academy of Artificial Intelligence (BAAI), the Natural Science Foundation of Guangdong Province (Grant No. 2021A1515012640), Basic Research Fund of Shenzhen City (Grant No. JCYJ20210324120012033 and JCYJ20190813165003837), and Overseas Cooperation Research Fund of Tsinghua Shenzhen International Graduate School  (Grant No. HW2021008). Finally, we would like to thank Luying Huang for valuable suggestions.

\bibliography{anthology,custom}
\bibliographystyle{acl_natbib}

\clearpage
\appendix
\section{Appendices}

\subsection{Pseudo-code of \MethodName{}}
\label{Appendix_A}
\begin{figure}[h]
\centering
\includegraphics[width=1.00\columnwidth]{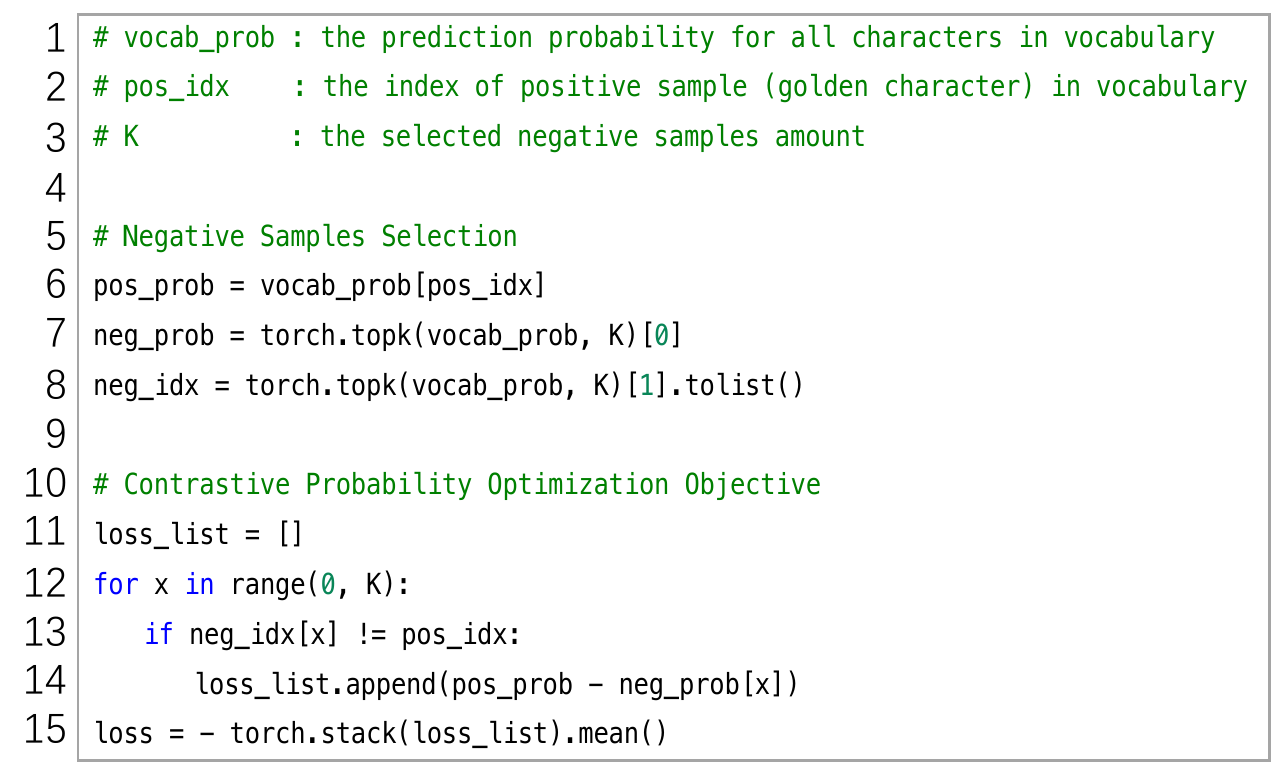}
\caption{Pseudo-code of our practical implementation.}
\label{Code_Figure}
\end{figure}

Figure~\ref{Code_Figure} shows the Pytorch-style pseudo-code for the \MethodName{}. As described in Section~\ref{Methodology}, our proposed \MethodName{} consists of two stages, namely Negative Samples Selection and Contrastive Probability Optimization. It is worthy noting that in the pseudo-code, we only show the process of calculating the loss of one training sample.

\subsection{Datasets Details}
\label{Appendix_B}
Table~\ref{Data_Statistics} shows the detailed statistics of our used datasets.  We report the number of sentences in the datasets (\#Sent), the average sentence length of the datasets (Avg.Length), and the number of misspellings the datasets contains (\#Errors).

\begin{table}[h]
\small
\centering
\begin{tabular}{lrrr}
\hline Training Data & \#Sent & Avg. Length & \#Errors \\
\hline SIGHAN13 & 700 & $41.8$ & 343 \\
SIGHAN14 & 3,437 & $49.6$ & 5,122 \\
SIGHAN15 & 2,338 & $31.3$ & 3,037 \\
Wang271K & 271,329 & $42.6$ & 381,962 \\
\hline Total & 277,804 & $42.6$ & 390464 \\
\hline \hline Test Data & \#Sent & Avg. Length & \#Errors \\
\hline SIGHAN13 & 1,000 & $74.3$ & 1,224 \\
SIGHAN14 & 1,062 & $50.0$ & 771 \\
SIGHAN15 & 1,100 & $30.6$ & 703 \\
\hline Total & 3,162 & $50.9$ & 2,698 \\
\hline
\end{tabular}

\caption{Statistics of the datasets that we use in experiments. All the training data are merged to train the models in our experiments. The test sets are used separately to evaluate performance.}
\label{Data_Statistics}
\end{table}

\subsection{Implementation Details }
\label{Appendix_D}
All the source code of our experiments is implemented using Pytorch~\cite{paszke2019pytorch} based on the Huggingface's implementation of Transformer library\footnote{https://github.com/huggingface/transformers}~\cite{wolf-etal-2020-transformers}. 
The architecture of the BERT encoder we use in the related models is same as the $BERT_{BASE}$ model, which has 12 transformers layers with 12 attention heads and its hidden state size is 768. We initialize the BERT encoder with the weights of Chinese BERT-wwm model~\cite{cui-etal-2020-revisiting}.
We train \MethodName{} with the AdamW~\cite{loshchilov2018fixing} optimizer for 10 epochs. The training batch size $N$ is set to 64 and the evaluation batch size is set to 50. The negative samples size $K$ is set to 5 by default. The weighting factors $\lambda_1 $, $ \lambda_2$ are both set to 1. In all our experiments, when $\lambda_1$ is equal to 1, it means that we use a fine-tuned BERT on the training set as the initialization model to continue the corresponding training process under different loss functions. The initial learning rate is set to 5e-5. We set the maximum sentence length to 128. The model is trained with learning rate warming up and linear decay.

It is worth noting that the annotation quality of SIGHAN13 test dataset is relatively poor. As we have observed and mentioned in~\cite{cheng-etal-2020-spellgcn,xu-etal-2021-read}, quite lots of the mixed usage of auxiliary (such as “\begin{CJK*}{UTF8}{gbsn}的\end{CJK*}”, “\begin{CJK*}{UTF8}{gbsn}地\end{CJK*}”, and “\begin{CJK*}{UTF8}{gbsn}得\end{CJK*}”) don't have correct annotations. Therefore, the evaluation metrics we use may not accurately reflect the real model performance on SIGHAN13. To alleviate this problem,  there are two main solutions in previous works.~\citet{cheng-etal-2020-spellgcn} propose to continue fine-tuning well-trained models on the SIGHAN13 training dataset before testing, which we think will suffer from the over-fitting problem. Therefore, we follow the post-processing method proposed in~\cite{xu-etal-2021-read} and don't consider all the detected and corrected mixed auxiliary. This approach does not compromise the fairness of the evaluation process and can better reflect the model performance.

\end{document}